\documentclass[a4paper, 10pt, conference]{ieeeconf}      % Use this line for a4 paper

\IEEEoverridecommandlockouts                              % This command is only needed if 
\usepackage{graphicx} % for pdf, bitmapped graphics files
\usepackage{amsmath} % assumes amsmath package installed

\usepackage{amsthm}
\usepackage{listings}
\theoremstyle{definition}

\usepackage{amssymb}
\usepackage{pifont}% http://ctan.org/pkg/pifont
\usepackage{booktabs}

\let\labelindent\relax
\usepackage{enumitem}
\usepackage[normalem]{ulem}

\usepackage{algorithm, algpseudocode}
\usepackage[colorlinks=true]{hyperref}
\usepackage[font=scriptsize,labelfont=bf]{caption}
\usepackage{etoolbox}
\makeatletter
\patchcmd{\@makecaption}
  {\scshape}
  {}
  {}
  {}
\makeatother
\title{\LARGE \bf
Work in Progress - 
Automated Generation of Robotic Planning Domains from Observations
}

\author{Maximilian Diehl and Karinne Ramirez-Amaro % stops a space
\thanks{Maximilian Diehl and Karinne Ramirez-Amaro. Faculty of Electrical Engineering, Chalmers University of Technology, SE-412 96 Gothenburg, Sweden
        {\tt\small \{diehlm, karinne\}@chalmers.se}}%
}

\begin{document}

\maketitle
\thispagestyle{empty}
\pagestyle{empty}

%%%%%%%%%%%%%%%%%%%%%%%%%%%%%%%%%%%%%%%%%%%%%%%%%%%%%%%%%%%%%%%%%%%%%%%%%%%%%%%%
\begin{abstract} 
%Move 1: Background/Introducation/Situation \\
%Move 2: Present research / purpose \\
%Move 3: Methods/Materials/Ssubjects/Procedure \\
%Move 4: Results / Findings \\
%Move 5: Discussion/conclusion/implications/recommendations \\ \\
In this paper, we report the results of our latest work on the automated generation of planning operators from human demonstrations, and we present some of our future research ideas. To automatically generate planning operators, our system segments and recognizes different observed actions from human demonstrations. We then proposed an automatic extraction method to detect the relevant preconditions and effects from these demonstrations. Finally, our system generates the associated planning operators and finds a sequence of actions that satisfies a user-defined goal using a symbolic planner. The plan is deployed on a simulated TIAGo robot. 
Our future research directions include learning from and explaining execution failures and detecting cause-effect relationships between demonstrated hand activities and their consequences on the robot's environment.
The former is crucial for trust-based and efficient human-robot collaboration and the latter for learning in realistic and dynamic environments.

\end{abstract}

 \section{\textsc{Introduction}}
%%%%%%%%%%%%%%%%%%%%%%%%%%%%%%%%%%%%%%%%%%%%%%%%%%%%%%%%%%
This paper presents preliminary results of our work on the automated generation of planning operators from human demonstrations.
The context of our approach is Automated Planning (AP), which is considered to be an essential tool in achieving the deliberation of autonomous robots~\cite{INGRAND201710}. AP is the process of planning ahead on how to reach a goal based on a set of available and applicable high-level actions~\cite{Arora2018}.

Generally, a planning task is split into \texttt{domain} and \texttt{problem}.
The \texttt{problem} defines the initial state and a goal of the planning problem. The \texttt{domain} contains a list of all possible actions, also called operators, that the system could use to progress from the initial state to the goal. Such operators are defined via name, a set of preconditions, and a set of effects. The \texttt{domain} is typically constructed by hand, which is very time-consuming and requires the expertise of domain experts~\cite{jimenez2012review, Jilani2014}.

%  \textbf{This paragraph looks incomplete, what is the take-away message? E.g. "However, generating this information is a ....." or "Nevertheless, this information is typically provided by an expert, thus the automatic generation needs to be investigated"}

% \begin{figure}[ht!]
% \centering
%   \includegraphics[width=0.44\textwidth]{./Images/MainFig.png}
%   \caption{This Figure captures the main idea of our latest approach~\cite{Diehl2021}. A domain file is automatically generated based on planning operators that are extracted from human demonstrations. This domain can be used to generate plans, for example, for new stacking goals, which are then executed in a simulation environment with the TIAGo robot.}
% \label{fig:overview}
% \end{figure}
\begin{figure}[ht!]
    \centering
      \includegraphics[width=0.48\textwidth]{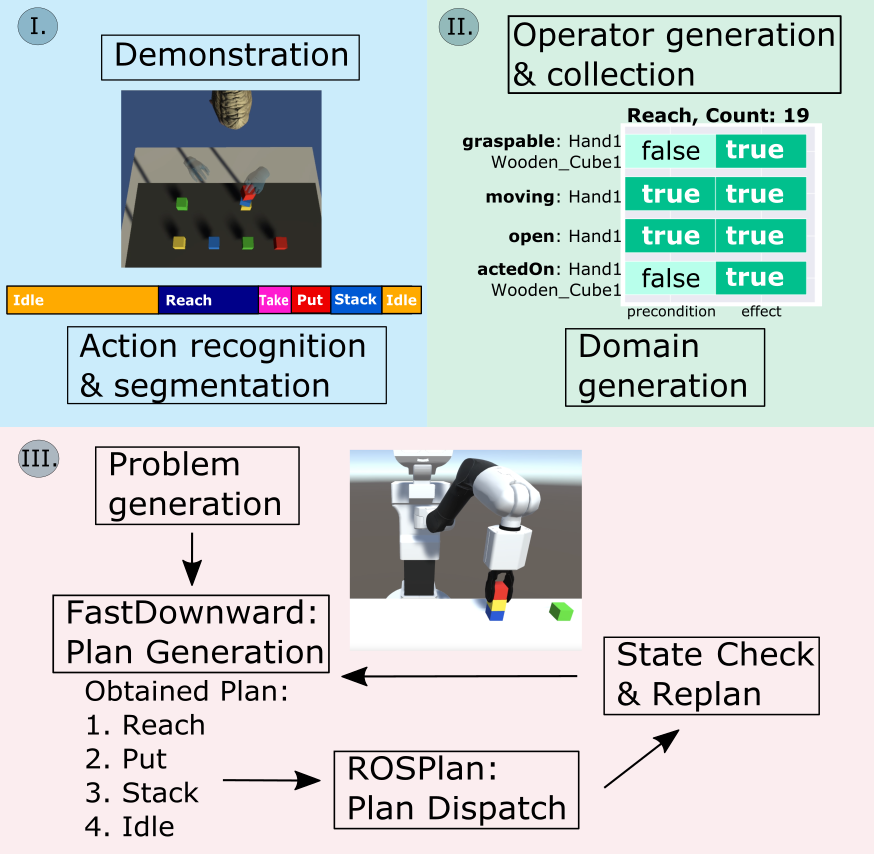}
      \caption{This figure illustrates the individual blocks of our system. I. The domain actions are obtained from human demonstrations performed in Virtual Reality. II. For the operator generation, actions are classified, and their preconditions and effects are extracted. At run-time, a planning domain is generated from the operator list. III. The specific planning problem comes from a user-defined goal state. The plan is constructed with the help of off-the-shelf symbolic planners like Fast-Downward and executed by the TIAGo robot in a simulated environment.}
    \label{fig:pipeline}
\end{figure}

We have, therefore, addressed the problem of \texttt{domain} construction, in our current work~\cite{Diehl2021}, by developing a novel method to automatically generate planning operators from human demonstrations in Virtual Reality (VR), visualized in Figure~\ref{fig:pipeline}. Our primary focus is to provide an action-oriented knowledge structure that collects high-level operators from new demonstrations continuously. Furthermore, the operators should generalize over object types and from human to robot, such that the agent can utilize already collected knowledge to plan even for previously unseen goals. The functionality and results of our previous work are elaborated in more detail in Section~\ref{sec:prevWork}.

Moving forward, we want to increase the applicability of our system by adding more complicated demonstration scenarios. In particular, we want to enable our system to handle timely shifted effects, parallel activities (e.g., demonstrations with two hands simultaneously), and erroneous effects. The key to solving these kinds of problems could be causality and causal inference, as it is introduced in \cite{Pearl09}.

In the light of the increasing potential of human-robot co-existence and collaboration, also the explainability of the agents' actions and failures is an important problem. 
Failures can occur on various levels of the planning and execution chain~\cite{Diehl2020, Bauer2020, Mitrevski2020}. Some works like~\cite{Mitrevski2020, Kaelbling2017, Agostini2020} try to learn parametrized abstract operators of lower-level actions, like pushing or pulling, which increases generalisability over different robotic platforms and failure diagnosis capabilities. Also, causality could help with this objective by providing links to the parameters on which a robot bases its decision. Ultimately, working on this problem motivates us to advance from deterministic to probabilistic operators and connect robot-specific diagnoses regarding things like collisions or other deviations from a planned trajectory with an operator refinement process, as further explained in Section~\ref{sec:workInProgress}.

\section{\textsc{Our Current System}}
%\section{\textsc{Work in Progress}}
\label{sec:prevWork}
%%%%%%%%%%%%%%%%%%%%%%%%%%%%%%%%%%%%%%%%%%%%%%%%%%%%%%%%%%
The pipeline of our current system is visualized in Figure~\ref{fig:pipeline}. The first step is the demonstration's activity segmentation, based on a decision tree as proposed in~\cite{ramirez17AIJ}. This is how we retrieve the names for the new operators. Beyond that, we implemented a method that automatically retrieves the relevant preconditions and effects and generates the new planning operators. Preconditions and effects are grounded predicates that can be either true or false and an operator would change one or several of these predicates.
One example of an operator with the name \texttt{Put}\footnote{Our definition of \texttt{Put} is equivalent to a \texttt{PickUp} action.}, obtained from our experiments, looks as follows:
\vspace{-1mm}
\begin{align*} %Put1, Count:
\footnotesize
    \text{Put}&(\texttt{Hand}, \texttt{Table}, \texttt{Cube}): \\
    &\text{\textbf{preconditions}}:  \\
        &{\tt inTouch}(\texttt{Cube},  \texttt{Table}), &{\tt onTop}(\texttt{Cube},  \texttt{Table}), \\ 
        &{\tt inHand}(\texttt{Hand}, \texttt{Cube}), &\neg {\tt handOpen}(\texttt{Hand}), \\
        &\neg {\tt handMove}(\texttt{Hand}) \\ 
        &\text{\textbf{effects}}: \\
        &\neg {\tt inTouch}(\texttt{Cube}, \texttt{Table}),
        &\neg {\tt onTop}(\texttt{Cube}, \texttt{Table}),\\ 
        &{\tt inHand}(\texttt{Hand}, \texttt{Cube}),
        &\neg {\tt handOpen}(\texttt{Hand}), \\
        &{\tt handMove}(\texttt{Hand}) 
\end{align*}
\noindent The \texttt{Put} operator has as input arguments a hand, a table, and a cube. It can only be used when the cube is \texttt{onTop} of and \texttt{inTouch} with the table but has already been grabbed by the hand; as an effect, the cube is lifted into the air, therefore \texttt{onTop} of and \texttt{inTouch} got negated, and the hand is in a moving state. Using these sets of conditions and effects, a symbolic planner can find a sequence of operators (and associated motor policies $\pi$) that can be applied in a new environment.

The main challenges that we addressed in our work~\cite{Diehl2021} are noisy or incomplete action demonstrations, filtering of operator irrelevant objects and relations, and generalization from specific objects (e.g.  \texttt{Right\_hand},  \texttt{Cube\_green1}) to generic types (e.g. \textit{Hand}, \textit{Wooden\_cube}). 

During each additional demonstration, newly observed operators are either added, in case they are unlike any other operator, in terms of preconditions and effects, or their count is incremented if already observed before. We also proposed to use the operator count as a prioritization measure during the planning process. This measure is based on the intuition that more commonly demonstrated actions are, first of all, more common and second, in the light of potential errors during the classification, also more robust. This allows us to build up and reuse knowledge even in new situations.

There are several ways to formulate a planning problem, and PDDL~\cite{pddl} is one of the most widespread ones. Supported by the popularity of the International Planning Competitions (IPC), a large variety of off-the-shelf planners have been and are continuously developed that support PDDL as input. We therefore wrote a parser that automatically collects the operator information into a \texttt{problem.pddl} file. Additionally, the \texttt{problem.pddl} file is automatically generated, which requires connection to the execution environment, and a manual goal state definition handled through a ROS service.

One problem with classic PDDL operators is that they are deterministic, even if the actual execution of the plan might not work as expected. We, therefore, connected our system with the ROSPlan framework~\cite{ROSPlan} and use the Esterel plan dispatcher for the plan dispatch. We set up the ROSPlan sensing interface \cite{Canal2019}, which allows the ROSPlan knowledge base to keep track of the current world state and to make sure that the required preconditions for each action are fulfilled before it is scheduled. Currently, we are also working on triggering a replanning call in case there is a discrepancy between the expected state and the actual state after each action execution.
% Even though replanning is popular~\cite{Yoon2007}, other work points out that replanning every time a failure occurs is not practical in real world robotic interaction, in particular for long and complex tasks. Therefore, reactive low-level controllers~\cite{Paxton2019RLDS} and Behavior Trees~\cite{Colledanchise2019} are presented.

%%%%%%%%%%%%%%%%%%%%%%%%%%%%%%%%%%%%%%%%%%%%%%%%%%%%%%%%%%
\section{Results}
%%%%%%%%%%%%%%%%%%%%%%%%%%%%%%%%%%%%%%%%%%%%%%%%%%%%%%%%%%

\begin{figure}[ht!]
    \centering
      \includegraphics[width=0.45\textwidth]{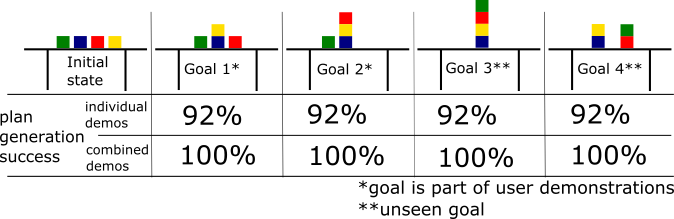}
      \caption{Plan goals with corresponding plan generation success ratio.}
    \label{fig:plangoals}
\end{figure}

For the evaluation of our system, we study the task of stacking cubes, which is a well-known problem within the robotics as well as planning community~\cite{AI}.
% \cite{Karapinar2015, DeLaCruz2020}
We recorded three participants to stack four different cube combinations each, including the stacking of one and two cubes with the left and right hand. We generated 12 domains based on the individual demonstration and one domain that combines all operators from the 12 separate demonstrations together. Based on all of these domains, plans for four different goals were generated. We observed that the plan generation and execution were successful for 11 out of the 12 individual demonstrations (92\%, Fig. \ref{fig:plangoals}, second row). Furthermore, we observed that for using all the 115 operators obtained from all the demonstrations combined, plan generation and execution were successful in all cases (Fig. \ref{fig:plangoals}, third row).

%For this experiment, we hand-coded the execution functions for each planned action and assumed them always to succeed. 
%To allow more robust action executions, we integrated ROSPlan into the execution pipeline. The sensing environment~\cite{Canal2019} keeps track of the current state during execution and stops executing a new action in case its preconditions are not met. However, ROSPlan does not automatically instruct a replan if there is a discrepancy between the overall expected state and the actual state after one action has finished. We tested this behavior with Goal 4 (stacking of two small towers) and added a small offset in the stacking goal location, which could be the result of wrong sensor information regarding the location of \texttt{cube\_blue} and \texttt{cube\_red}. Consequently, \texttt{cube\_green} and \texttt{cube\_yellow} were not stacked on top but next to the cubes on the table. Nevertheless, the plan execution moved on because all of the preconditions for the next actions were satisfied. With the replanning module, however, we can take care of these situations and make sure the actual goal is met.

%%%%%%%%%%%%%%%%%%%%%%%%%%%%%%%%%%%%%%%%%%%%%%%%%%%%%%%%%%
%\section{\textsc{Outlook}}
\section{\textsc{Work in Progress}}
\label{sec:workInProgress}
%%%%%%%%%%%%%%%%%%%%%%%%%%%%%%%%%%%%%%%%%%%%%%%%%%%%%%%%%%
In this section, we discuss several aspects that attracted our attention during the development of the system that we have introduced in Section~\ref{sec:prevWork} and that we want to investigate in the future.
%%%%%%%%%%%%%%%%%%%%%%%%%%%
\subsection{Learning cause and effect relationships}
%%%%%%%%%%%%%%%%%%%%%%%%%%%
An interesting point that came up during the operator generation process's design concerns the question of cause and effect. The human demonstrator, through its hands, or the robot with its gripper, manipulates the environment. 
%Hand activities can or can not result in a change in the environment. 
However, not every hand activity results in an environmental change. Two operators generated from the demonstrations were, for example, the \texttt{Reach} and \texttt{Put} operator. The main effect of the \texttt{Reach} operator is that an object is graspable, e.g., the right hand is reaching for a distant cube which is graspable as a result. Ideally, the reaching process was without collisions, which means that the environment has not seen any perceivable changes through the reaching execution. The \texttt{Put} operator, on the other hand, describes an activity where the cube is lifted from the table into the air. In this case, not just the robot state but the environment has changed, in particular, \texttt{onTop} and \texttt{inTouch} are negated throughout the operator execution. From a human perspective, the cause for the cube not being on top of the table anymore is, without doubt, the hand activity of picking the cube up and putting it somewhere. 

In our current cube stacking experiment setup, we are dealing with quite immediate cause-effect relationships. In reality, however, there are many possibilities in which the automatic and reliable recognition of what exactly caused a change in the environment is less straightforward. Consider the following examples:  
\begin{enumerate}
    \item \textit{Timely shifted effects}: Performing a hand activity leads to an environmental effect that is not immediately perceivable. An example could be turning on the stove to gradually heat it up.
    \item \textit{Parallel activities:} During our experiments, we explicitly instructed the participants only to use one hand to stack the cubes. In reality, however, we commonly use both hands in parallel. In these cases, it is more difficult to assign environment effects to the correct hand activity automatically. Imagine the following example: A human picks up a bottle of water from the table with his right hand while reaching towards the bottle with the left hand, with the intention to eventually unscrew the bottle. During this process, the \texttt{onTop} predicate, describing the relationship between the bottle and the table, turns from true to false. Both hands perform an activity that involves the bottle, but nevertheless, for the human, it is instantly clear that it was the right hand that was responsible for this change since it had the bottle \texttt{inHand}. The robot, however, first needs to learn these causal relations.
    % \item \textit{External environment changes}: How can we discern environment changes that are not caused by the agent itself, but, either by external factors, like the blast of wind that blew a pile of papers from the desk, or even by other agents? %\textbf{Do we really want o investigate this?}
    \item \textit{Erroneous effects}: Last but not least, it is very important to learn and understand the cause for failures. Imagine a situation where the robot would stack cubes, but does not place them centered on top of each other. After the third cube, the tower falls. The robot should be able to understand, that the issue was the incorrect positioning of (some) cubes, refine the operators and be able to explain what went wrong. 
\end{enumerate}
The concept of cause and effect is viewed as a fundamental challenge in Artificial Intelligence~\cite{Pearl09, Scholkopf2021}
%\cite{Bhattacharjya2020, Scholkopf2021}
, but even though planning operators are inherently causal, causality has not been widely explored in existing operator generation methods \cite{Arora2018, Scholkopf2021}. This leads us to our first research question:

% \textcolor{red}{The concept of cause and effect is viewed as a fundamental challenge in Artificial Intelligence~\cite{Pearl09, Scholkopf2021}but is usually not fully considered in previous work about planning operator generation \cite{Pasula2004}. Popular measures are the assumption that environments do not change other than through interactions by the agent, or alternating observation sources, like kinesthetic teaching on the robot itself or non-parallel plan traces. This leads us to our first research question:}
\begin{itemize}[align=left]
    \item[\textbf{RQ1:}] How can we reliably detect cause-effect relationships in human demonstrations involving timely shifted, erroneous effects, and parallel hand activities, for the learning of planning operators?
\end{itemize}

% \textcolor{red}{As a starting point, we are considering the transition from deterministic to probabilistic operators since the robot will have several precondition/effect possibilities with varying probabilities for each operator. With more experience, either through additional demonstrations, or self-exploration in a simulation environment, the robot can then refine the probabilities, thus resulting in convergence towards the true operator configuration.}

The first step is going to be to identify causal relations in our demonstrations. In particular, we would like to detect causal relations between predicates, e.g., between robot predicates like \texttt{inHand} and environment predicates like \texttt{onTop}. In the example of parallel hand activities, this would help us better differentiate which hand is responsible for which environment change. One of the most popular algorithms for detecting causality in data is the PC-algorithm~\cite{Spirtes2000}. It is based on conditional independence tests but can, in certain instances, only retrieve the skeleton of the resulting graph without any directions.
On the other hand, interventional methods allow for perfect reconstruction of the causal relationships, including directions \cite{Eberhardt2006}.  We could retrieve this kind of data either through interventional instructions so that the demonstrator performs interventions based on the interventional pattern known or even determined by the robot. The second option is to enrich observational data through self-exploration on the robot side, either in a simulation or directly in the real world.
%Retrieving causal relations will help us design better operator generation algorithms that ground the decision on which predicates to consider on causal connections. 

% As a start point we are considering probabilistic planning operators~\cite{Martinez2015, Bauer2020}. Due to 

% but more sophisticated algorithms need to be developed in order to take a step from modeled lab-examples to complicated real-world scenarios that might incorporate some or all of the three points mentioned above.

% As a start point we are starting probab. operators, 

%%%%%%%%%%%%%%%%%%%%%%%%%%%
%\subsection{Learning from failures, explaining failures, generalization of actions}
\subsection{Learning from and explaining failures}
%%%%%%%%%%%%%%%%%%%%%%%%%%%
%Works like~\cite{Bauer2020} bring up another important point. 
In~\cite{Bauer2020} the authors argue that learning operator effects in complex real-world scenarios have to be learned over time, thus, enabling robots to \textit{learn from failures}.
One and the same action performed on/with different objects might have different consequences, which depend on the objects' properties. The specific example investigated in~\cite{Bauer2020} is dropping items of different sizes into a bowl. Some things, like the football, are too large to fit, and as a result, dropping a football does not lead to the same effect as dropping a table tennis ball. In this way, they connect knowledge about objects captured in an ontology into the planning process to generalize actions over objects reliably.
% Questions: is the cube stacking scenario good enough to show something similar? Where can we take this? How can we exploit our previous work for that?

From \textit{learning from failures} it is only a small step to another crucial objective of trust-based human-robot co-existence and collaboration: \textit{Explaining failures}. 
Recent work \cite{Kaelbling2017, Mitrevski2020} investigates how symbolic operators can capture the essence of low-level actions, which has the advantage of action generalisability over different robotic platforms and increases failure diagnosis capabilities. In~\cite{Kaelbling2017} a parameterized model of a pushing activity is learned, which relates the applied force to the final block location after sliding. The authors of~\cite{Mitrevski2020} abstract tasks from the sensorimotor level to state variables like {\tt inFrontOf}, {\tt behind}, or {\tt above}. The task of opening a drawer and a door is learned based on these spatial and temporal features.

Considering the examples of related work mentioned in this section, it becomes clear that failures can occur on different levels, including low-level (execution) failures, path planning failures (e.g., collisions), and high-level planning failures (e.g., due to incorrect operators). This leads us to our second research question: 
\begin{itemize}[align=left]
    \item[\textbf{RQ2:}] How can we detect and explain execution failures such that we can address refinements at the correct level of the planning-execution chain?
\end{itemize}
There are several extensions we want to incorporate into our current system to tackle RQ2. Our current main concern is the refinement of erroneous operators. However, given the close interplay between the different layers, investigating them entirely separately might not be possible. The success of an operator execution through the robot might depend on the specific situation and the particular objects involved. For example, the success chance to stack two balls is lower than stacking two cubes. Even the specific embodiment of the robot, e.g., kind of gripper or motor accuracy, might play a role and will differ from the human.
As a starting point, we, therefore, need to consider the transition from deterministic to probabilistic operators since the robot will have several precondition/effect possibilities with varying probabilities for each operator. Causal inference could again be a helpful tool to increase generalizability between different situations or objects. Coming back to our cube stacking example: a causal analysis could reveal which and how strong some of the object properties, like shape, size, and color, are responsible for the stacking failures. This kind of information could be used for operator refinement and is helpful for the explainability component of plan execution.
Instead of reusing erroneous operators, they could be stored in a separate domain of erroneous operators, further investigated once the robot is not required for an important task. To connect high-level errors with lower-level errors, we then want to investigate the interplay between the operator and its execution in more detail. In other words, we need to connect irregularities like deviations from the planned path or collisions with the known chances of operator failures. This could allow us to refine either transition probabilities or add new effects altogether, but also decide when a refinement is not required. Ultimately, it will also be helpful to expand our set of predicates to be able to tackle more complex scenarios. That will be beneficial for both RQs we posed. One possibility could be to adopt the set of object-centered predicates proposed in \cite{Agostini2020}.

\section{\textsc{Conclusion}}
%%%%%%%%%%%%%%%%%%%%%%%%%%%%%%%%%%%%%%%%%%%%%%%%%%%%%%%%%%
This paper presents a system for the automated generation of planning domains, allowing robots to accomplish new tasks based on human demonstrations. Results show that as much as one single stacking demonstration can be enough (92\%) to allow for successful plan generation and execution of various unseen stacking scenarios. The system also allows for continuous skill collection and prioritizes more often observed operators based on the operator cost minimization during the planning process. 
%Our action-centered skill acquisition approach shows great potential for collecting, storing, refining and combining knowledge for autonomous robots. 

Furthermore, we elaborated on the lessons learned from the development of our current system and formulated interesting future directions. 
Our next goal is the investigation of \textit{cause and effect} relationships in the context of planning operator generation and \textit{learning from and explaining failures} occurring on different levels of the planning and execution chain.

\addtolength{\textheight}{-12cm}   % This command serves to balance the column lengths
                                  % on the last page of the document manually. It shortens
                                  % the textheight of the last page by a suitable amount.
                                  % This command does not take effect until the next page
                                  % so it should come on the page before the last. Make
                                  % sure that you do not shorten the textheight too much.

%%%%%%%%%%%%%%%%%%%%%%%%%%%%%%%%%%%%%%%%%%%%%%%%%%%%%%%%%%%%%%%%%%%%%%%%%%%%%%%%

%%%%%%%%%%%%%%%%%%%%%%%%%%%%%%%%%%%%%%%%%%%%%%%%%%%%%%%%%%%%%%%%%%%%%%%%%%%%%%%%

%%%%%%%%%%%%%%%%%%%%%%%%%%%%%%%%%%%%%%%%%%%%%%%%%%%%%%%%%%%%%%%%%%%%%%%%%%%%%%%%

% Appendixes should appear before the acknowledgment.

\section*{ACKNOWLEDGMENT}
The research reported in this paper has been supported by Chalmers AI Research Centre (CHAIR).

\bibliography{mybib}
\bibliographystyle{IEEEtran}

\end{document}